%% file: main.tex
\def\isarxiv{1} %%%Arxiv version, we uncomment this line

\ifdefined\isarxiv
\documentclass[11pt]{article}
\else
 
\documentclass[conference]{IEEEtran}
\IEEEoverridecommandlockouts
\usepackage{hyperref}

\def\BibTeX{{\rm B\kern-.05em{\sc i\kern-.025em b}\kern-.08em
    T\kern-.1667em\lower.7ex\hbox{E}\kern-.125emX}}
\fi

\usepackage[utf8]{inputenc}

\usepackage{microtype}
\usepackage{amsmath}
\usepackage{amsthm}
\allowdisplaybreaks
\usepackage{amssymb}
\usepackage{algorithm}
\usepackage{subfig}
\usepackage{color}
\usepackage{epstopdf}
\usepackage{url}
\usepackage{graphicx}
\usepackage{color}
\usepackage{epstopdf}
\usepackage{algpseudocode}
\usepackage{scrextend}
\usepackage{bbm}
\usepackage{comment}

\usepackage{multirow}
\usepackage{dsfont}
\usepackage{mathtools}
\usepackage{enumitem}

\usepackage[makeroom]{cancel}
\usepackage{stmaryrd}
\usepackage{booktabs, makecell}
\usepackage{pifont}% http://ctan.org/pkg/pifont
\usepackage{lipsum}

\usepackage{amsfonts}       % blackboard math symbols
\usepackage{nicefrac}       % compact symbols for 1/2, etc.
\usepackage{xcolor}         % colors

\usepackage{tikz}
\usepackage{hyperref}
\definecolor{mydarkblue}{rgb}{0,0.08,0.45}
\hypersetup{colorlinks=true, citecolor=mydarkblue,linkcolor=mydarkblue}

\usetikzlibrary{arrows}
\ifdefined\isarxiv
\usepackage[margin=1in]{geometry}
\else

\fi
\graphicspath{{./figs/}}

\definecolor{b2}{RGB}{51,153,255}
\definecolor{mygreen}{RGB}{80,180,0}
\definecolor{yl}{RGB}{255,80,0}
\definecolor{myl}{RGB}{180,80,20}

\newtheorem{theorem}{Theorem}[section]
\newtheorem{lemma}[theorem]{Lemma}
\newtheorem{definition}[theorem]{Definition}

\newtheorem{proposition}[theorem]{Proposition}
\newtheorem{corollary}[theorem]{Corollary}

\newtheorem{assumption}[theorem]{Assumption}

\newtheorem{remark}[theorem]{Remark}

\renewcommand{\tilde}{\widetilde}

\newcommand{\wt}{\widetilde}

\DeclareMathOperator{\R}{{\mathbb R}}

\DeclareMathOperator*{\E}{{\mathbb{E}}}

\newcommand{\fedavg}{{\texttt{FedAvg}}} % Macro for the method

\ifdefined\isarxiv
\title{A Convergence Theory for Federated Average: Beyond Smoothness\thanks{A preliminary version of this paper appeared in BigData'2022.}}

\date{}

\author{
Xiaoxiao Li\thanks{
\texttt{xiaoxiao.li@ece.ubc.ca}. University of British Columbia.}
\and 
Zhao Song\thanks{\texttt{zsong@adobe.com}. Adobe Research.}
\and 
Runzhou Tao\thanks{\texttt{runzhou.tao@columbia.edu}. Columbia University.}
\and 
Guangyi Zhang\thanks{
\texttt{guangyi.zhang@mail.mcgill.ca}. McGill University.}
}

\else 
\title{A Convergence Theory for Federated Average: Beyond Smoothness} 
\author{\IEEEauthorblockN{Xiaoxiao Li}
\IEEEauthorblockA{\textit{Department of ECE} \\
\textit{UBC}\\
Vancouver, Canada \\
xiaoxiao.li@ece.ubc.ca
}
\and
\IEEEauthorblockN{Zhao Song}
\IEEEauthorblockA{
\textit{Adobe Research}\\
\textit{Adobe}\\
San Jose, USA \\
zsong@adobe.com}
\and
\IEEEauthorblockN{Runzhou Tao}
\IEEEauthorblockA{\textit{Department of CS} \\
\textit{Columbia University}\\
New York, USA \\
runzhou.tao@columbia.edu}
\and
\IEEEauthorblockN{Guangyi Zhang}
\IEEEauthorblockA{\textit{Department of ECE} \\
\textit{McGill University}\\
Montreal, Canada \\
guangyi.zhang@mail.mcgill.ca}
}
 
\fi

\begin{document}

%\ifdefined\isicml

\ifdefined\isarxiv

\begin{titlepage}
\maketitle
\begin{abstract}

\input{abstract}

\end{abstract}
\thispagestyle{empty}
\end{titlepage}

\else

\maketitle
\begin{abstract}
Federated learning enables a large amount of edge computing devices to learn a model without data sharing jointly. As a leading algorithm in this setting, Federated Average (\fedavg), which runs Stochastic Gradient Descent (SGD) in parallel on local devices and averages the sequences only once in a while, have been widely used due to their simplicity and low communication cost. However, despite recent research efforts, it lacks theoretical analysis under assumptions beyond smoothness. In this paper, we analyze the convergence of {\fedavg}. Different from the existing work, we relax the assumption of strong smoothness.  More specifically, we assume the semi-smoothness and semi-Lipschitz properties for the loss function, which have an additional first-order term in assumption definitions. In addition, we also assume bound on the gradient, which is weaker than the commonly used bounded gradient assumption in the convergence analysis scheme. As a solution, this paper provides a theoretical convergence study on Federated Learning. 
\end{abstract}

\begin{IEEEkeywords}
Federated learning, semi-smoothness, no-critical-point, semi-Lipschitz.
\end{IEEEkeywords}
\fi

\input{intro}   %%% Section 1 Introduction
\input{related} %%% Section 2 Related Work
\input{prob}    %%% Section 3 Problem
\input{result}  %%% Section 4 Our Result
\input{proofsketch} %%% Section 5 Proof Sketch
\input{problem} %%% Section 6 proof of our optimization
\input{proof_sketch_app} %%% Section 7 Supplementary proof
\input{discussion}  %%% Section 8 Discussion

\ifdefined\isarxiv 
\bibliographystyle{alpha}
\else 
\bibliographystyle{IEEEtran}
\fi
\bibliography{ref}

\end{document}

%% file: abstract.tex
Federated learning enables a large amount of edge computing devices to learn a model without data sharing jointly. As a leading algorithm in this setting, Federated Average (\fedavg), which runs Stochastic Gradient Descent (SGD) in parallel on local devices and averages the sequences only once in a while, have been widely used due to their simplicity and low communication cost. However, despite recent research efforts, it lacks theoretical analysis under assumptions beyond smoothness. In this paper, we analyze the convergence of {\fedavg}. Different from the existing work, we relax the assumption of strong smoothness.  More specifically, we assume the semi-smoothness and semi-Lipschitz properties for the loss function, which have an additional first-order term in assumption definitions. In addition, we also assume bound on the gradient, which is weaker than the commonly used bounded gradient assumption in the convergence analysis scheme. As a solution, this paper provides a theoretical convergence study on Federated Learning.

%% file: intro.tex
\section{Introduction}

With the growing of computational power on edge devices, such as mobile phones, wearable devices, smart watches, self-driving cars, and so on, developing distributed optimization methods to address the needs of those applications is increasingly demanded. There are three core challenges existing in the distributed computing applications, including expensive communication, privacy concerns, and heterogeneity. To tackle the above-mentioned challenges, federated learning (FL) has emerged as an important paradigm in today's machine learning for distributed learning that enables different clients (also known as nodes) to collaboratively learn a model while keeping their private data. 

To train an FL algorithm in a distributed manner, the clients must transmit their training parameters to a central server. Typically, the central server has the same model architecture as the local clients. 
Similar to centralized parallel optimization, FL lets the clients do most of the computation while the central server updates the model parameters using the descending directions returned by the local clients. 

However, learning with FL significantly differs from the traditional parallel optimization in distributed learning in the various needs, including piracy requirements, large-scale machine learning and  efficiency. To meet these unique requirements, the most popular existing and easiest to implement FL strategy is Federated Average (\fedavg) \cite{mmr+17}, where clients collaboratively send updates of locally trained models to a global server. Each client runs a local copy of the global model on its local data. The global model's weights are then updated with an average of local clients' updates and deployed back to the clients. This strategy builds upon previous distributed learning work by supplying local models and performing training locally on each device. Hence \fedavg\ potentially empowers clients (especially clients with small datasets) to collaboratively learn a shared prediction model while keeping all training data locally. 

Although \fedavg\ has shown successes in classical Federated Learning tasks, it suffers from slow convergence and low accuracy in most non-iid contents \cite{lsz+20,lhy+19}.  There have been many efforts developing convergence guarantees for FL algorithms, i.e., how the convergence rate is affected by the client local update epochs, how many communication rounds are required to achieve a targeted model performance.  There have been many efforts developing convergence guarantees for FL algorithms \cite{kmr19,yyz19,wts+19,kkm+20} on \fedavg.

\textbf{Yet}, the optimization and convergence analysis in FL is \textit{quite non-trivial}. On the one hand, the optimized objective are usually not only non-convex but even non-smooth. For example, mapping functions may have non-linear operations, such as ReLU activation and maxing out the label for objective functions. On the other hand, different from centralized training, each client's local update steps and the difference between local and global model (aka \textit{local drift}) need to be considered.

In this paper, we study the following fundamental question:
\begin{quote}
    \begin{center}
{\it 
Can we show that \fedavg{} converges under mild assumptions? } 
\end{center}
\end{quote}

To answer the above question, the following two challenges need to be addressed: 1) How to find suitable assumptions and 2) How to design a framework to handle local updates and local drift. 
Inspired by overparameterized optimization theory, we introduce the semi-smooth and no-critical-point properties.
We develop a new framework to reason about the convergence of federated learning.
Our proof solves the key task of bounding the local drift under the semi-smoothness assumption.  

\paragraph{Our Contributions} We summarize our technical contributions below. 

\begin{itemize}
  \item We propose a new theoretical framework for federated learning under semi-smoothness settings. All the previous federated learning results either require smoothness or Lipschitz. However, our results make milder assumptions. We also relax the strong bounded gradient assumption popularly used at earlier convergence analyses.
    \item Local drift is a bottleneck in the convergence of FL training algorithms. Under our valid assumptions, the local drift in \fedavg{} is appropriately bounded when the local updating learning rate is controlled by the parameters associated with the assumptions.
    \item Our theoretical results indicate how local updates steps and learning rate affect the number of communication rounds in \fedavg.
\end{itemize}

\paragraph{Organization.} The paper organization is as follows. In Section~\ref{sec:related_work}, we discuss related work about FL algorithms, especially variants of \fedavg, and its convergence analysis under different settings. In Section~\ref{sec:problem}, we describe our problem setting that we consider \fedavg{} algorithm and state the semi-smoothness, no-critical-point and semi-Lipschitz assumptions to be used in our proof. In Section~\ref{sec:framework}, we elaborate on the rationality of making the three assumptions and state our results on convergence of \fedavg\ under these assumptions. In Section~\ref{sec:proof_sketch}, we give a proof sketch of our results first on a simplified non-FL case, and then on the FL case. In Section~\ref{sec:discussion}, we provide a conclusion for the paper and discuss possible future work and the social impact of this work.

%% file: related.tex
\section{Related Work}\label{sec:related_work}

\paragraph{Federated learning} 
With the growth of computational power, data are massively distributed over an incredibly large number of devices. Federated learning is proposed to allow machine learning models to be trained on local clients in a distributed fashion.
An essential bottleneck in such a distributed training on the cloud is the communication cost. 

Federated average (\fedavg)~\cite{mmr+17} firstly addressed the communication efficiency problem. \fedavg\ algorithm allows devices to perform local training of multiple epochs to reduces the number of communication rounds, then average model parameters from the client devices.  Later, a myriad of variations and adaptations have arisen~\cite{wys+20,zll18,kma+19,ljz+21,hlsy21,dms+21,syz22_iclr,dmszl21,swyz22}.

As stated earlier, federated learning involves learning a centralized model from distributed client data. This centralized model benefits from all client data and can often
result in a beneficial performance e.g. in including next word prediction \cite{hrm+18, yae+18}, emoji prediction \cite{rmr19+}, vocabulary estimation \cite{cmo+19}, and predictive models in health \cite{lgd+20, lst+20}. Multiple research efforts studying the issues on more efficient communication strategies \cite{kmrr16,bdkd20,krsj19,yhwz+19,hkmm20}. Existing studies on federated learning have mostly focused on improving communication efficiency, understanding the effect of sampling a subset of clients in each round of communication, and heterogeneous data distribution.

\paragraph{Convergence of \fedavg}  For identical clients, \fedavg\ coincides with
parallel SGD analyzed by \cite{zwls10} who proved asymptotic convergence. \cite{s18} and, more recently \cite{sk19, pd19,kmr20}, gave a sharper analysis of the same method, under the name of local
SGD, also for identical functions. The analysis of \fedavg\ is more sophisticated than parallel SGD due to local drift, which represents the difference between local and global models. The divergence is empirically observed in \cite{zll18} on non-iid data. Some analyses constrain this drift by assuming a bounded gradient~\cite{wts+19,yyz19}. Although a few recent studies~\cite{ghr21,klb+20} do not require bounded gradient, their theories are limited to L-smoothness  assumption. Specifically, \cite{ghr21} only discusses the convergence on convex cases and ~\cite{klb+20} relax the bound to grow with gradient norm for non-convex cases. Alternatively \cite{kmr20} treat the drift as additional noise.
In recent work, \cite{kkm+20} proposes to reduce the gradient diversity, where authors suggest augmenting the local gradients with a controlled variance.

%% file: prob.tex
\section{Preliminary}\label{sec:problem}

In this section, we introduce the setting of our problem. We first introduce the notations to be used throughout the paper. Then, we formulate our learning model by defining the local and total loss function and explaining our \fedavg algorithm. Finally we state the three non-smooth assumptions, namely semi-smoothness, no-critical-point and semi-Lipschitz, based on which we prove \fedavg's convergence.

\paragraph{Notations}
For any positive integer $n$, we use $[n]$ to denote set $\{1,2,\cdots,n\}$. 
For a vector $x$, we use $\| x \|_2$ to denote its $\ell_2$ norm. For a matrix $W$, we use $\| W \|$ to denote the spectral norm of $W$. We use $\| W \|_F$ to denote its Frobenius norm. We use $\E[\cdot]$ to denote the expectation of a random variable if its expectation is existing. We use $\Pr[]$ to denote the probability.

Let $n$ denote the number of input data points.
Let $N$ denote the number of clients. We can think of each client will have $n/N$ data points.
Let $S_1 \cup S_2 \cup \cdots \cup S_N = [n]$ and $S_i \cap S_j = \emptyset$.
Given $n$ input data points and labels 
\begin{align*}
\{ (x_1,y_1), (x_2,y_2), \cdots, (x_n,y_n) \} \in \R^d \times \R.
\end{align*}

\paragraph{Problem formulation} In this work, we consider the following federated learning model using \fedavg\ algorithm. Suppose $N$ clients are in the federated learning system.  We define the local loss function $L_c$ of $c$-th client for $c \in [N]$,
\begin{align*}
    L_c (W,x) = ~ \frac{1}{2} \sum_{i \in S_c} \text{loss}( x_i, y_i )^2,
\end{align*}
where $c$-th client holds training data $\{(x_i,y_i) ~|~ i \in S_c\}$, \text{loss} can be $l_2$ loss, cross entropy loss, and others in practice. When all clients are activate, the total loss function is defined as
\begin{align*}
    L(W,x) =   \frac{1}{N} \sum_{c=1}^N  L_c(W,x).
\end{align*}
We formalize the problem as minimizing the sum of loss functions over all clients:
\begin{align*}
    \min _{W \in \mathbb{R}^{d \times M}}
    L(W).
\end{align*}
We define 
\begin{align*}
g_{c}(W):=\nabla L_{c} (W ; \zeta_c )
\end{align*}

be an unbiased stochastic gradient of $L_c$ with variance bounded by $\sigma^2$.
\paragraph{Algorithm} A typical implementation of \fedavg\ contains an additional global model (with the same architecture as a local model) and performs in the following way. First, the local client (say the $c$-th) trains the local neural network and updates local model weight $W_c$. Then, the local model weight $W_c$ is sent to the global model. Later, the global model average the received local model weights and broadcast the averaged weights as $U$ to local models for client model updating. We depict the pseudo-code of \fedavg\ algorithm in Algorithm~\ref{ag1}. It worth noting that, different from centralized training, \fedavg\ allows clients to update $K$ epochs before aggregation locally. Such design has been shown to reduce the communication round but induces challenges of analyzing the convergence of \fedavg{}.

To tackle the convergence problem of \fedavg{}, we propose the following assumptions. The justifications are detailed in Section~\ref{sec:framework}.
\begin{assumption}
\label{assump:1}
We state the assumptions
\begin{itemize}
\item Semi-smoothness, Section~\ref{sec:smoothness} 
\begin{align*}
    L(W) \leq L(U) + \langle \nabla L(U), W- U \rangle 
     +  b \| U - W \|^2 + a \| U - W \| \cdot L(U)^{1/2}.
\end{align*}

\item No critical point, Section~\ref{sec:no_critical_point}
\begin{align*}
    \tau_1^2 L(U) \leq \| \nabla L(U) \|^2 \leq \tau_2^2 L(U).
\end{align*}
 
\item Semi-Lipschitz, Section~\ref{sec:lipschitz}
\begin{align*}
    \|\nabla L (W ) - \nabla L(U)\|^2  \leq & ~ \beta^2  \|W - U\|^2+  \alpha^2 \|W - U\|L(U)^{1/2}.
\end{align*}
\end{itemize}
\end{assumption}

\section{Framework Going Beyond Smoothness}\label{sec:framework}

In this section, we give formal definitions of the three conditions we use to prove the convergence of \fedavg. We relax the smoothness condition in the classical analysis to semi-smoothness which is shown to be held by neural networks. No critical point condition weakens the bounded gradient property that are widely used in previous convergence analysis on federated learning algorithms. And the semi-Lipschitz condition weakens the classical Lipschitz property.

\subsection{Smoothness property}\label{sec:smoothness}
To ensure the objective function decreases over training time, one relies on the smoothness property in classical optimization theory. We first start with describing the definition of $\beta$-smoothness in FL, which is extended from the classical analysis.
\begin{definition}[$\beta$-smoothness]
\label{def:smooth}
For any function $L$, we say it is $\beta$-smooth if for any $W,U$
\begin{align*}
    L(U) \leq L(W) + \langle \nabla L(W), U-W \rangle + \frac{\beta}{2} \|W-U\|^2.
\end{align*}
\end{definition}
 
However, the neural networks, the widely used model in FL, may not meet the twice differentiablity requirement of $\beta$-smoothness (i.e., the ReLU activation). Thus, a milder assumption of smoothness is often required. To deal with the issue, in \cite{als19_dnn}, semi-smoothness is proposed and shown to be held by neural networks. To extend the semi-smoothness definition in FL, we have:
\begin{definition}[$(\alpha,\beta)$-semi-smoothness]
\label{def:semi-smooth}
For any function $L$, we say it is $(\alpha,\beta)$-semi-smooth if for any $W, U$ 
\begin{align*}
    L(U) \leq & ~ L(W) + \langle \nabla L(W), U- W \rangle 
     +
    b \| W - U \|^2   
    +  a \| W - U \| \cdot L(W)^{1/2}. 
\end{align*}

\end{definition}
It is worth noting that, different from the smoothness definition, we have an additional first order term $\| W - U \|$ on the right hand side.

\subsection{No critical point}\label{sec:no_critical_point}
Finding approximate critical points of a non-smooth and non-convex function was challenging\cite{blo05}, until \cite{als19_dnn} proof that the gradient bounds for points that are sufficiently close to the random initialization. It is proved in \cite{als19_dnn} that there is no critical point for square loss function for neural networks. Thus the no critical point (Theorem 3 in  \cite{als19_dnn}) is a property in the nature of neural networks in the classical training regime. Therefore, we extend the definition of the no critical point property to FL as:

\begin{definition}[No critical point] 
\label{def:ncp}
Let $\mathcal{U}$ be a neighbor set of $U^*$ (a minimum of $L$). We say there is no critical point for a function $L$, there exist constants $0 <\tau_1 < 1$ and $\tau_2 >0 $, if for any $U \in \mathcal{U}$
\begin{align*}
    \tau_1^2 \cdot L(U) \leq \|\nabla L(U)\|^2 \leq \tau_2^2 \cdot L(U).
\end{align*}
\end{definition}
 
Definition \ref{def:ncp} shows that the gradient norm is large when the objective function is large. This means that there are no saddle points or critical points when we are sufficiently close to the random initialization. Thus, we hold a good brief of finding global minima of the objective function.

Bounded gradient property is a popular scheme for convergence analysis used in some of the previous work~\cite{ly17,yyz19}, which is  defined as:
\begin{definition}[Bounded gradient \cite{lhy+19}]
We say a function has bounded gradient, if there exists $G \geq 0$ such that for any $U$, $\|\nabla L(U)\|^2 \leq G^2$ holds.
\end{definition}
 
The no critical point assumption is a weaker assumption for both strong convexity and bounded gradient.  

\subsection{Lipschitz property}\label{sec:lipschitz}
In the classical analysis, the $\beta$-Lipschitz defined in Definition \ref{def:lipschitz} is assumed. We also have the $\beta$-smoothness defined in Definition \ref{def:smooth} is implied by the $\beta$-Lipschitz. 

\begin{definition}[$\beta$-Lipschitz]
\label{def:lipschitz}
For a function $L$, we say it is $\beta$ smooth if for any $W,U$
\begin{align*}
    \| \nabla L (W) - \nabla L(U) \| \leq \beta \cdot \| W - U \|.
\end{align*}
\end{definition}
\cite{als19_dnn} shows that for the overparameterized neural networks, the first-order term is much smaller than the second-order term during neural networks evolution. In this case, Definition~\ref{def:semi-smooth} ($(\alpha,\beta)$-semi-smoothness) is close to, but still not interchangeable with the classical Lipschitz smoothness.
We propose to consider a weaker definition of Lipschitz in FL, due to the multiple local client update steps. The milder assumption is defined as $(\alpha,\beta)$-semi-Lipschitz.
\begin{definition}[$(\alpha,\beta)$-semi-Lipschitz]
\label{def:semi_lipschitz}
For a function $L$, we say it is $(\alpha,\beta)$-semi-smoothness if for any $W,U$
\begin{align*}
   \| \nabla L (W) - \nabla L(U) \|^2 \leq \beta^2 \cdot \| W - U \|^2 
    + \alpha^2 \|W-U\| \cdot \max \{ L(W)^{1/2}, L(U)^{1/2} \}.
\end{align*}

\end{definition}
Since the semi-Lipschitz definition should be symmetric in terms of $U$ and $W$, we use 
\begin{align*} 
 \max \{ L(W)^{1/2}, L(U)^{1/2} \}
 \end{align*}

instead of $L(W)^{1/2}$ as in Definition \ref{def:semi-smooth}.

%% file: result.tex
\subsection{Our results}

Standard global convergence analysis for gradient descent uses smoothness and strong convexity. We show the relaxed assumptions, $(a,b)$-semi-smoothness and $(\tau_1, \tau_2)$-no critical point, are sufficient to analyze the convergence for gradient descent.   
To derive the convergence for {\fedavg}, we also need to bound local gradient updates, which requires the semi-Lipschitz property. 

Let $U^*$ denote a minimizer of $L$. Under Assumption~\ref{def:semi-smooth}, \ref{def:ncp} and \ref{def:semi_lipschitz}, we state our result and provide a proof sketch (see Section~\ref{sec:main1_informal}).
 
\begin{theorem}[Main result]
\label{thm:main1_informal}
If the loss functions $L_c$ ($c \in [N]$) for each client $c$ is:    $(a,b)$-semi-smooth, $(\alpha,\beta)$-semi-Lipschitiz, $(\tau_1,\tau_2)$-non-critical point, 
for any parameters $\eta_l$ less than
\begin{align*} 
  \min\{1/(\alpha^2 K), 1/(10^2K\tau_2(\beta+\alpha)), 1/(10^2\sqrt{K} (\beta + \alpha) ) \} 
 \end{align*} 
 and
 \begin{align*} 
 \eta_g \leq \tau_1^2/(20Kb\eta_l),
\end{align*}
 
for {\fedavg}, we have 
\begin{align*}
    & ~ \E[L(U^r) - L(U^*)] 
    \leq  (1-\lambda_1)^r \cdot (L(U^0) - L(U^*)) + 2\lambda_2,
\end{align*}
where 
\begin{align*}
\lambda_1 = & ~ \frac{K \eta_l \eta_g}{4}(1-4bK \eta_l \eta_g - 2a)\tau_1^2 \\
\lambda_2 = & ~ ( 1 + a + b K \eta_l \eta_g) \frac{K \eta_l \eta_g}{10} \sigma^2 .
\end{align*}
\end{theorem}

\begin{corollary}\label{coro:main1}
For any desired $\epsilon$, using a step-size of
\begin{align*}
    \eta_g \leq \min\{ \tau_1^2/(20Kb\eta_l), 2 \epsilon / (\sigma^2(1+a+\tau_1^2/20K)) \},
\end{align*}
after rounds $    R = \log (\frac{2 (L(U^0) - L(U^*))}{\epsilon}) \frac{1}{\lambda_1}$,  we have that 
 \begin{align*} 
\E[L(U^r) - L(U^*)] \leq \epsilon.
\end{align*}
\end{corollary}
Given the definition of $\lambda_1$, we notice the trade-off that either a small or a large $K$ will results a large $R$ under the assumptions.
 

%% file: proofsketch.tex
\section{Proof Sketch}\label{sec:proof_sketch}
In this section, we show a proof sketch of our main result Theorem~\ref{thm:main1_informal}. To better illustrate the proof structure, we first prove the convergence of the non-federated learning gradient descent algorithm under the semi-smoothness and non-critical-point conditions. The simplified proof shares the same structure as the final proof. Then, we consider the federated learning case with four steps, including the key step of bounding local drift, and discuss the difference between FL and non-FL cases. Due to page limit, this proof sketch uses gradient descent instead of stochastic gradient descent and the full proof is given in Section~\ref{sec:proof_theorem1}.

\subsection{Non-federated learning case (simplified)}
\begin{proposition}
\label{prop:non_fed}
Let $x^*$ denotes a minimum of $L$. Suppose we run gradient descent algorithm to update $x_{t+1}$ in each iteration as follows:
 \begin{align*}
    x_{t+1} = x_t - \eta \cdot \nabla L(x) |_{x = x_t}.
 \end{align*}
If the loss function $L$, $(a,b)$-semi-smooth, $(\tau_1,\tau_2)$-non-critical point, $0.5\tau_1^2 \geq a \tau_2$, 
using $\eta \le \tau_1^2/(10 b \tau_2^2)$ then we have
 
\begin{align*}
    L(x_{t+1}) - L(x^*) \leq (1-\lambda ) (L(x_t) - L(x^*))
\end{align*}
where $\lambda = 0.1 \eta \tau_1^2$.
\end{proposition}
\begin{proof}
We delay the proof to Section~\ref{sec:proof_non_fed_app}.
\end{proof}

\subsection{Federated learning case, Theorem~\ref{thm:main1_informal}}\label{sec:main1_informal}

In this section, we show a proof sketch for Theorem \ref{thm:main1_informal}. We show the simplified proof with gradient descent. The complete proof with stochastic gradient descent is presented in Section \ref{sec:proof_theorem1}.

The same as first step Eq.~\eqref{eq:non_fed1} in the proof of Proposition \ref{prop:non_fed}, the first step in our proof is using the semi-smoothness of $L$ to compute
\begin{align}
\label{eq:sketch1}
  L(U^{r+1}) - L(U^*) 
 \leq  L(U^{r}) - L(U^*) + \langle \nabla L(U^{r}), \Delta U^{r} \rangle + b\|\Delta U^{r}\|^2 + a\|\Delta U^{r}\| \cdot L(U^{r})^{1/2}
\end{align}
The gradient update $\Delta U^{r}$ is an average of local gradients. Unlike the centralized case, the terms $\langle \nabla L(U^{r}), \Delta U^{r} \rangle$ and $\|\Delta U^{r}\|$ cannot be simply bounded by $(\tau_1. \tau_2)$-non-critical point assumption. 

Next, we are going to show the bounds for the RHS. The remained proof is organized as follows:  
\begin{description}
    \item[Bounding $\langle \nabla L(U^r), \Delta U^r \rangle$.]
    
    In the non-federated case, we have 
    \begin{align*} 
    \langle \nabla L(U^r), \Delta U^r \rangle = -\wt \eta \|\nabla L(U^r)\|^2.
    \end{align*}
    In the federated case, $\Delta U^{r}$ is an average of local gradients, but we need an upper bound for $\langle \nabla L(U^r), \Delta U^r \rangle$ in terms of global parameter $U^{r}$ rather than local parameters $W^{r}_{c,k}$. Thus, we upper bound $\langle \nabla L(U^r), \Delta U^r \rangle$ by \begin{align*}
    -0.5\wt \eta \|\nabla L(U^r)\|^2
    \end{align*}
    plus differences of gradients of loss at local and global parameters, i.e. 
    \begin{align*}
    \|\nabla L_c(W^r_{c,k-1}) -\nabla L_c(U^r) \|^2.
    \end{align*}
    
    To bound 
    \begin{align*}
    \|\nabla L_c(W^r_{c,k-1}) -\nabla L_c(U^r) \|^2,
    \end{align*}
    we need the Lipschitz property, which is not used in the non-federated case. We define 
\begin{align}
\label{eq:def_xi}
    \xi = \frac{1}{KN}\sum_{k=1}^{K}\sum_{c=1}^{N}\|W_{c,k-1}-U\|^2.
\end{align} 
    Then, $\langle \nabla L(U^r), \Delta U^r \rangle$ is upper bound by a sum of $-0.5\wt \eta \|\nabla L(U^r)\|^2$, $\xi$ and $L(U^r)$.
    
    \item[Bounding $\| \Delta U^r \|^2$.]
    
    The same as the non-federated case, we use non-critical point property to bound the norm of gradients, i.e. for all $k \in [K]$, \begin{align*} \|W^r_{c,k}\|^2 \leq \tau_2^2 L(W^r_{c,k}).
    \end{align*}
    Then we use the result of the non-federated case, for all $k \in [K]$, 
    \begin{align*}
    L(W^r_{c,k}) \leq L(U^r).
    \end{align*}
    
    \item[Bounding $\xi$.]
    
    The term $\xi$ is an average of local drifts caused by the local updates. It is expected that this drift is small for overparameterized neural network. We upper bound $\xi$ by induction. 
    
    \item[Choosing parameters.] Finally, we put all the bounds together and explain how to choose parameters to assure decrease in loss for each global training round.
\end{description}

We consider gradient descent in this proof sketch for the ease of presentation, the proof for stochastic gradient descent is shown in the Section~\ref{sec:proof_non_fed_app}. The gradient update for {\fedavg} is 
\begin{align*}
    U^{r+1} & = U^{r} + \Delta U^{r} \\
     & = U^{r} - \frac{\tilde{\eta}}{K N} \sum_{k \in [K], c \in [N]} \nabla L_{c}\left(W^{r}_{c, k-1}\right),
\end{align*}
where $W_{c,k}^{r}$ is defined by
 \begin{align*}
     W_{c,0}^{r}  = U^{r}, \;\;\;
     W_{c,k}^{r}  = W_{c,k-1}^{r} - \eta_l \nabla L_c(W_{c,k-1}^{r}),
 \end{align*}
where $\wt \eta = K \eta_g \eta_l$ is the effective step.

\paragraph{Bounding $\langle \nabla L(U^{r}), \Delta U^{r} \rangle$}
To simplify the notation, we ignore the superscript $r$. First, we compute the term $\langle \nabla L(U), \Delta U \rangle$ in Eq.~\eqref{eq:sketch1}.
{%\small
\begin{align}
 \langle \nabla L(U), \Delta U \rangle 
=  & ~ -\tilde{\eta} \langle \nabla L(U), \frac{1}{K N} \sum_{k \in [K], c \in [N]} \nabla L_{c}\left(W_{c, k-1}\right) \rangle \notag \\
    \leq & ~ -\frac{\tilde{\eta}}{2} \|\nabla L(U)\|^2  + \frac{\tilde{\eta}}{2} \Big\| \frac{1}{KN} \sum_{k \in K, c\in [N]} \nabla L_c(W_{c,k-1}) -\nabla L(U) \Big\|^2 \nonumber\\
    = & ~ -\frac{\tilde{\eta}}{2} \|\nabla L(U)\|^2  +  \frac{\tilde{\eta}}{2} \Big\| \frac{1}{KN} \sum_{k \in K, c\in [N]} ( \nabla L_c(W_{c,k-1}) -\nabla L_c(U) ) \Big\|^2 \nonumber\\
    \leq & ~ -\frac{\tilde{\eta}}{2} \|\nabla L(U)\|^2  +  \frac{\tilde{\eta}}{2KN} \sum_{k \in K, c\in [N]} \|\nabla L_c(W_{c,k-1}) -\nabla L_c(U) \|^2, \label{eq:sketch2}
\end{align}
}
where second step follows from 
$-ab =  \frac{1}{2}((b-a)^2-a^2) -\frac{1}{2}b^2 
\leq  \frac{1}{2}((b-a)^2-a^2)$, 
the third step follows from 
$\nabla L(U) = \frac{1}{KN}\sum_{k \in [K],c \in [N]} L_c(U)$,
and the last step follows from  
$(\sum_{i=1}^n a_i)^2 \leq n\sum_{i=1}^n a_i^2$.

To bound the difference of gradients of loss, one have to use the Lipschitz property  as discussed in section \ref{sec:lipschitz}. 
By $(\alpha, \beta)$-semi-Lipschitz and $2ab \leq a^2+b^2$, we have
\begin{align}
\label{eq:sketch4}
    \|\nabla L_c(W_{c,k-1}) -\nabla L_c(U) \|^2  
 \leq & ~  \beta^2  \|W_{c,k-1} - U\|^2  
 + \alpha^2 \|W_{c,k-1} - U\|L_c(U)^{1/2} \notag \\
    \leq & ~ (\beta^2+\frac{\alpha^2}{2}) \|W_{c,k-1} - U\|^2 + \frac{\alpha^2}{2} L_c(U)
\end{align}

Combining the definition of $\xi$ in \eqref{eq:def_xi} and Eq.~\eqref{eq:sketch2}, \eqref{eq:sketch4}, we have 
\begin{align}
\label{eq:sketch5}
 \langle \nabla L(U), \Delta U \rangle 
    \leq & ~ -\frac{\tilde{\eta}}{2} \|\nabla L(U)\|^2 + \frac{2\beta^2 + \alpha^2}{4} \wt \eta \cdot \xi 
    +  \frac{\alpha^2}{4} \wt \eta \cdot L(U) \notag \\
    \leq & ~ - \frac{2\tau_1^2 - \alpha^2}{4} \wt \eta \cdot L(U) + \frac{2\beta^2 + \alpha^2}{4} \wt \eta \cdot \xi,
\end{align}
where the second step follows from the non-critical point property.

\paragraph{Bounding $\|\Delta U\|^2$}
Next, we consider the term $\|\Delta U\|^2$ in Eq.~\eqref{eq:sketch1},
\begin{align}
    \label{eq:sketch6}
    \|\Delta U\|^2 
\leq & ~ \frac{\tilde{\eta}^2}{KN}\sum_{k \in [K], c \in [N]}\|\nabla L_c(W_{c,k-1})\|^2 \notag \\
\leq & ~ \frac{\tau_2^2\tilde{\eta}^2}{KN}\sum_{k \in [K], c \in [N]} L_c(W_{c,k-1}),
\end{align}
where the first step follows from 
 \begin{align*}
(\sum_{i=1}^n a_i)^2 \leq n\sum_{i=1}^n a_i^2,
 \end{align*}
and the second step follows from the non-critical point property.

By applying Proposition \ref{prop:non_fed} to $L_c(W_{c,k-1})$, for any positive integer $k \geq 2$, we have
\begin{align}
\label{eq:sketch7}
    L_c(W_{c,k}) \leq L_c(W_{c,k-1}) \leq \cdots \leq L_c(U) 
\end{align}

Combining Eq.~\eqref{eq:sketch6} and \eqref{eq:sketch7}, $\|\Delta U\|^2$ is upper bounded as
\begin{align}\label{eq:sketch8}
    \|\Delta U\|^2 
\leq \tau_2^2\tilde{\eta}^2 L(U).
\end{align}

\paragraph{Bounding $\xi$}
Then, we upper bound $\xi$ by induction. For $K=1$, $\xi=0$, and for $k \geq 1$, we compute one step local update. 
\begin{align*}
\|W_{c,k+1} - U\|^2 
=  & ~  \|W_{c,k} - U - \eta_l \cdot \nabla L_{c} (W_{c,k}) \|^{2} \notag \\
\leq & ~ (1 + \frac{1}{K-1} ) \cdot \|W_{c,k}-U \|^{2} +K \eta_l^{2} \|\nabla L_{c} (W_{c,k} )\|^{2}  \notag \\
\leq & ~ (1 + \frac{2}{K} ) \cdot \|W_{c,k}-U \|^{2} +K \eta^{2} \|\nabla L_{c} (W_{c,k} )\|^{2}  \notag \\
\leq & ~ (1 + \frac{2}{K} ) \cdot \|W_{c,k}-U \|^{2} + KG^2 \eta^{2} L_c(W_{c,k})  \notag \\
\leq & ~ (1 + \frac{2}{K} ) \cdot \|W_{c,k}-U \|^{2} + K\tau_2^2 \eta_l^{2} L_c(U),
\end{align*}
where the second step follows from 
$2ab \leq t\cdot a^2+ \frac{1}{t} \cdot b^2$, the third step follows from $K \geq 2$, and the last step follows from the non-critical point property, and Eq.~\eqref{eq:sketch7}.

Unrolling the recursion above,
\begin{align*}
     \E[\|W_{c,k} - U\|^2] 
    \leq & ~ (K\tau_2^2 \eta_l^{2} L_c(U)) \cdot \sum_{i=1}^{K} (1+\frac{2}{K})^{i-1} \notag \\
    \leq & ~ 4K^2\tau_2^2 \eta_l^{2} \cdot L_c(U),
\end{align*}
where the geometric sequence
 \begin{align*}
\sum_{i=1}^{K} (1+\frac{2}{K})^{i-1} \leq 4K.
 \end{align*}
Averaging over $c$ and $k$, we upper bound 
\begin{align}
\label{eq:sketch9}
    \xi  
    \leq 4 K^2 G^2 \eta_l^2 \frac{1}{N}\sum_{c=1}^{N} L_c(U) 
   = 4 K^2 G^2 \eta_l^2 L(U).
\end{align}

\paragraph{Putting together}
By combining Eq.~\eqref{eq:sketch1}, \eqref{eq:sketch5}, \eqref{eq:sketch8}, and \eqref{eq:sketch9}, we have 
\begin{align}
\label{eq:sketch10}
L(U^{r+1}) - L(U^*)
 \leq & ~ L(U^{r}) - L(U^*) - \frac{2\tau_1^2 - \alpha^2}{4} \wt \eta \cdot L(U^{r}) \notag\\
 + & ~ (2\beta^2 + \alpha^2) K^2 G^2 \wt \eta \eta_l^2 \cdot L(U^{r}) \notag \\
 + & ~ b \tau_2^2\tilde{\eta}^2 \cdot L(U^{r}) + a \tau_2 \tilde{\eta} \cdot L(U^{r}) \notag \\
\leq & ~ (1-A)L(U^{r}) - L(U^*),
\end{align}
where
\begin{align*}
  A = & ~  \frac{2\tau_1^2 - \alpha^2}{4} \wt \eta - a \tau_2 \tilde{\eta} 
   - (2\beta^2 + \alpha^2) K^2 G^2  \eta_l^2 \wt \eta - b \tau_2^2\tilde{\eta}^2 .
\end{align*}
\paragraph{Choosing parameters}
We need to carefully tune the parameters to find a $\gamma \geq 0$, such that $A \geq \gamma$.
 
By choosing 
\begin{align*}
& ~ \alpha \leq  0.5 \tau_1 , \;\;
a\tau_2 \leq  0.1\tau_1^2 ,\\
& ~ \eta_l \leq  \tau_1/(4KG(2\beta+\alpha)) ,\;\;
\wt \eta \leq  \tau_1^2/(16b\tau_2^2),
\end{align*}
we have
\begin{align}
\label{eq:sketch11}
    A \geq \frac{\tau_1^2}{8}\wt \eta := \gamma.
\end{align}
By plugging Eq.~\eqref{eq:sketch11} into Eq.~\eqref{eq:sketch10}, we get
\begin{align*}
c L(U^{r+1}) - L(U^*) 
\leq  & ~ (1-\gamma)L(U^{r}) - L(U^*) \notag \\
\leq & ~ (1-\gamma)(L(U^{r}) - L(U^*)),
\end{align*}
where the second step follows from $L(U^*) \geq 0$.

\begin{remark}
To better present the flow of the proof, this proof sketch uses gradient descent instead of stochastic gradient descent and uses loose upper bound for $\|\Delta U\|^2$ and $\|\Delta U^{r}\| \cdot L(U^{r})^{1/2}$. In the Section \ref{sec:proof_theorem1}, we give complete proof and tighter upper bound so that the restrictions for parameters $\alpha$, $a$ and $\tau_2$ can be relaxed. 
\end{remark}

%% file: problem.tex
\begin{algorithm}[!ht]\caption{$\fedavg$ learning algorithm.}\label{ag1}
\begin{algorithmic}[1]
\State Initialize $U(1)$   
	\For{$r = 1 \to R$}
	    \State Clients run \textbf{Procedure A} in parallel
	    \State Server run \textbf{Procedure B}
	\EndFor
	\State \Return $U(R + 1)$
	\Statex
	\Procedure{\textbf{A}. ClientUpdate}{$r,c$} 
	         \State $W_c \leftarrow U$ 
        \For{$k=1,\ldots,K$} 
            \State $W_c \leftarrow W_c - \eta_{\mathrm{local}} \cdot g_c(W_c) $ 
        \EndFor
        \State 
        $\Delta W_c \leftarrow W_c - U$
	    \State Send $\Delta W_c(r)$ to \textsc{ServerExecution}
	\EndProcedure
 
    \Procedure{\textbf{B}. ServerExecution} {$r$}:
                \For{each client $c$ \textbf{in parallel}}
                \State  \Comment{Receive local model weights update}
                \State{$\Delta U_c(r) \gets$ \textsc{ClientUpdate}$(r,c)$}
               
  	        \State $\Delta U(r) \leftarrow \frac{1}{N} \sum_{c \in [N]}  \Delta U_c(r)$  
  	        \State \Comment{Aggregation on the server side.}
	        \State $U(r+1) \leftarrow U(r) + \eta_{\mathrm{global}} \cdot \Delta U(r)$ 
                \State{Send $U(r+1)$ to client $c$ for \textsc{ClientUpdate}$(r,c)$}
        \EndFor
        \EndProcedure
	
	\end{algorithmic}
\end{algorithm}

\section{Proof of our optimization}\label{sec:proof_theorem1}
\subsection{Separating mean and variance}
\begin{lemma}[separating mean and variance, Lemma 4 in \cite{ kkm+20}]
\label{lem:smv}
Let $a_{1}, \ldots, a_{n} $ be $n$ random variables in $\R^{d}$ which are not necessarily independent. First suppose that their mean is $\E[a_i] = \mu_i$ and variance is bounded as $\E[\|a_i-\mu_i\|^2] \leq \sigma^2$. Then, the following holds
\begin{align*}
    \E [ \|\sum_{i=1}^{n} a_{i} \|^{2} ] \leq \|\sum_{i=1}^{n} \mu_{i} \|^{2}+n^{2} \sigma^{2}.
\end{align*}
Now instead suppose that their conditional mean is $\E [a_{i} \mid a_{i-1}, \ldots a_{1} ]=\mu_{i}$, i.e. the variables $ \{a_{i}-\mu_{i} \}$ form a martingale
difference sequence, and the variance is bounded by $\E [ \|a_{i}-\mu_{i} \|^{2} ] \leq \sigma^{2}$ as before. Then we can show the tighter bound
\begin{align*}
    \E [ \|\sum_{i=1}^{n} a_{i} \|^{2} ]  \leq 2 \|\sum_{i=1}^{n} \mu_{i} \|^{2}+2 n \sigma^{2}.
\end{align*}
\end{lemma}

\subsection{Bounded drift}
We define the $\sigma$ as follows
\begin{definition}[variance]\label{def:sigma}
Let $S_c$ be a random sample of data in client $c$. Let $g_c(W) = \frac{\partial L_c(W;S_c)}{ \partial W }$. Then $\E_{S_c}[g_c(W)] = \frac{\partial L_c(W)}{ \partial W } = \nabla L_c(W)$. We define $\sigma$ to be
\begin{align*}
    \sigma := \max_{ c \in [N] } \Big( \E_{S_c} [ \| g_c(W) - \E_{S_c}[ g_c(W) ] \|^2 ] \Big)^{1/2}.
\end{align*}
\end{definition}

\begin{lemma}[bounded drift, a general version of Lemma 8 in \cite{kkm+20}]
\label{lem:xi_bound}
 Let $N$ denote the number of clients. Let $K$ denote the number of local updates. Let $\sigma$ be defined as Definition~\ref{def:sigma}. Let learning rate $\eta$ be satisfying that $\eta \in (0, \min \{ 1/(\alpha^2 K), 1/( 2 \beta K) \} )$. 

Assume the loss functions $L_c$ ($c \in [N]$) for each client $c$ is
\begin{itemize}
	\item $(\alpha,\beta)$-semi-Lipschitiz 
	\item $(\tau_1,\tau_2)$-non-critical point, 
\end{itemize}

We define $\xi$ as follows:
\begin{align*}
    \xi:=\frac{1}{K N} \sum_{k=1}^{K} \sum_{c=1}^{N} \E[\|W_{c,k}-U\|^{2}].
\end{align*}
 
We have
\begin{align*}
\xi  \leq 50 \eta^2 ( 1 + K^2 \cdot \tau_2^2 \cdot L(U) + K \cdot \sigma^2 ).
\end{align*}
\end{lemma}
\begin{proof}
If $K = 1$, the lemma holds since $W_{c,1} = U$ for all $c \in [N]$ and $\xi = 0$. 
Recall that the local update made on client $i$ is $W_{c,k+1} = W_{c,k} - \eta \cdot g_c(W_{c,k})$. Then,  {%\small
\begin{align}\label{eq:bound2_xi_eq1}
 \E [ \|W_{c,k+1} - U \|^{2} ] 
 = & ~ \E[ \|W_{c,k} - U - \eta \cdot g_{c} (W_{c,k} ) \|^{2}]  \notag \\
\leq & ~ \E[ \|W_{c,k} - U - \eta \cdot \nabla L_{c} (W_{c,k}) \|^{2}] + \eta^{2} \sigma^{2} 
\end{align}}
where the last step follows from Lemma \ref{lem:smv}.

For the first term in Eq.~\eqref{eq:bound2_xi_eq1}, we can upper bound it as follows:{%\small
\begin{align}\label{eq:bound2_xi_eq2}
& ~ \E[ \|W_{c,k} - U - \eta \cdot \nabla L_{c} (W_{c,k}) \|^{2}] \notag \\
\leq & ~  \E [\|W_{c,k}-U \|^{2}] + \eta^{2} \|\nabla L_{c} (W_{c,k} )\|^{2}  + 2 \E [\|W_{c,k}-U \|^{2} \cdot \eta \|\nabla L_{c} (W_{c,k} )\|]  \notag\\
\leq & ~ (1 + \frac{1}{K-1} ) \cdot \E [\|W_{c,k}-U \|^{2}] +K \eta^{2} \|\nabla L_{c} (W_{c,k} )\|^{2}  \notag \\
\leq & ~ (1 + \frac{2}{K} ) \cdot \E [\|W_{c,k}-U \|^{2}] +K \eta^{2} \|\nabla L_{c} (W_{c,k} )\|^{2}  \notag \\
\leq & ~ (1 + \frac{2}{K} ) \cdot \E [\|W_{c,k}-U \|^{2}] + 2K \eta^{2} \|\nabla L_{c} (W_{c,k} ) - \nabla L_{c}(U)\|^{2}   + 
2K \eta^2 \|\nabla L_{c}(U)\|^2  ,
\end{align}}
where the first step is because the triangle inequality,
 the second step follows from 
$2ab \leq \frac{1}{k-1}\cdot a^2+ (k-1) \cdot b^2$, the third step follows from $K \geq 2$, the last step follows from $(a+b)^2 \leq 2 a^2 + 2b^2$. 
 
For the second term in the Eq.~\eqref{eq:bound2_xi_eq2}, we can upper bound it as follows:
\begin{align}\label{eq:bound2_xi_eq3}
 2K \eta^{2} \|\nabla L_{c} (W_{c,k} ) - \nabla L_{c}(U)\|^{2}
\leq & ~  2 K \eta^2\cdot ( \beta^2 \E [\|W_{c,k}-U \|^{2}] + \alpha^2  \E[\|W_{c,k}-U\|] L_c(U)^{1/2} )  \notag \\
\leq & ~ \frac{1}{K} \cdot \E [\|W_{c,k}-U \|^{2}] + 2\alpha^2 K \eta^2 \E [\|W_{c,k}-U\|] L_c(U)^{1/2}   \notag \\
\leq & ~ \frac{1}{K} \cdot \E [\|W_{c,k}-U \|^{2}] + \frac{1}{K}  \cdot 2 \E [\|W_{c,k}-U\|] \cdot \eta L_c(U)^{1/2}  \notag \\
\leq & ~ \frac{2}{K} \cdot \E [ \| W_{c,k} - U \|^{2}] + \frac{1}{K} \cdot \eta^2 L_c(U) 
\end{align}
where 
the first step follows from $(\alpha,\beta)$-semi-Lipschitz, the second step follows from $\eta \leq \frac{1}{ 2 \beta K}$, the third step follows from $\eta \leq \frac{1}{ \alpha^2 K}$, the last step follows from $2ab \leq a^2+ b^2$, and $\eta \leq 1$.

Putting Eq.~\eqref{eq:bound2_xi_eq1}, \eqref{eq:bound2_xi_eq2}, \eqref{eq:bound2_xi_eq3} together, we have
{%\small
\begin{align*}
\E [ \|W_{c,k+1} - U \|^{2} ] 
\leq & ~ (1+\frac{4}{K}) \cdot \E [\|W_{c,k}-U \|^{2}] + \frac{1}{K} \eta^2 L_c(U) \\
& ~ +  2K \eta^2 \|\nabla L_{c}(U)\|^2 +\eta^{2} \sigma^{2} 
\end{align*}
}

For $k \geq 2$ and $K \geq 2$, unrolling the recursion above,
\begin{align}\label{eq:bound2_xi_eq4}
\E[\|W_{c,k} - U\|^2] 
     \leq & ~ \sum_{i=1}^{k-1} (\frac{1}{K} \eta^2 L_c(U) + 2K \eta^2 \|\nabla L_{c}(U)\|^2 +\eta^{2} \sigma^{2})(1+\frac{4}{K})^{i-1} \notag \\
    \leq & ~ (\frac{1}{K} \eta^2 L_c(U) + 2K \eta^2 \|\nabla L_{c}(U)\|^2 + \eta^2\sigma^{2}) \cdot \sum_{i=1}^{K} (1+\frac{4}{K})^{i-1}
\end{align}
We can bound the geometric sequence as follows:
\begin{align}\label{eq:bound2_xi_eq5}
    \sum_{i=1}^{K} (1+\frac{4}{K})^{i-1} 
    = & ~ 
    ((1+\frac{4}{K})^{K-1} - 1)\frac{K}{4} \notag \\
    \leq & ~ 
    (e^4-1)\frac{K}{4} \notag \\
    \leq & ~ 20 K,
\end{align}
where the second step follows from $(1+\frac{4}{n})^n \leq e^4$ for all positive integer $n$, and the last step follows from $(e^4-1)/4 \leq 20$.

Combining Eq.~\eqref{eq:bound2_xi_eq4} and Eq.~\eqref{eq:bound2_xi_eq5}, we have for any $k \leq K$, {%\small
\begin{align}\label{eq:bound2_xi_eq7}
 \E[\|W_{c,k} - U\|^2] \leq 20 K \cdot  (\frac{1}{K} \eta^2 L_c(U) + 2K \eta^2 \|\nabla L_{c}(U)\|^2 + \eta^2\sigma^{2}).
\end{align}}

Averaging over $c$ and $k$,
\begin{align*}
    \xi & = \frac{1}{K N} \sum_{k=1}^{K} \sum_{c=1}^{N} \E[\|W_{c,k}-U\|^{2}] \\
    & \leq 20\eta^2 \frac{1}{N}\sum_{c=1}^{N} L_c(U) + 40K^2 \eta^2 \cdot \frac{1}{N} \sum_{c=1}^{N}\|\nabla L_c(U)\|^2 +20K\eta^2\sigma^2 \\
    & \leq  (20 \eta^2 + 40 K^2 \eta^2 \tau_2^2) L(U) +20K\eta^2\sigma^2.
\end{align*}
where the the second step follows from Eq.~\eqref{eq:bound2_xi_eq7}, 
last step follows from the upper bound of no-critical point assumption.
\end{proof}

\subsection{One round progress}

Here we show a lemma for the one round bound for non-convex function.
\begin{lemma}[one round progress for non-convex function]
\label{lem:nonconvex_one_round}

Suppose the loss functions $L_c$ ($c \in [N]$) for each client $c$ is
\begin{itemize}
    \item $(a,b)$-semi-smooth,
	\item $(\alpha,\beta)$-semi-Lipschitiz.
\end{itemize}

We define $\xi$ as follows:
\begin{align*}
    \xi:=\frac{1}{K N} \sum_{k=1}^{K} \sum_{c=1}^{N} \E[\|W_{c,k}-U\|^{2}].
\end{align*}
We define effective step-size as $\tilde{\eta}:=K \eta_g \eta_l$ and server update in each round as:
\begin{align*}
\Delta U : =-\frac{\tilde{\eta}}{K N} \sum_{k \in [K], c \in [N]} g_{c}\left(W_{c, k-1}\right).
\end{align*} 
Then, for a tuning parameter $\eta$ and effective step-size $\tilde{\eta}$, the updates of \fedavg\ satisfy

\begin{align*}
    \E[L(U+\Delta U)] - L(U) 
    \leq  & ~ -\frac{\tilde{\eta}}{2}(1-4b\tilde{\eta} - \frac{2a\tilde{\eta}}{\eta}) \|\nabla L(U)\|^2 \\
    & ~ + \tilde{\eta}^2(2 \beta^2 + \alpha^2)(\frac{1}{2\tilde{\eta}}+b+\frac{a}{2\eta})\xi + \tilde{\eta}^2 (b+\frac{a}{2\eta}) \sigma^2 \\
     & ~ + \Big(\frac{\tilde{\eta}\alpha^2}{4} + \tilde{\eta}^2 \alpha^2 (b+\frac{a}{2\eta}) + \frac{a \eta}{2} \Big) \cdot L(U).
\end{align*}
\end{lemma}

\begin{proof}
By Assumption 1 in the lemma statement, we have
{%\small 
\begin{align}
\label{eq:nonconvex_one_round1}
   \E[L(U+\Delta U)] - L(U) 
    \leq & ~ \langle \nabla L(U), \E[ \Delta U ] \rangle + b \E[\|\Delta U\|^2] + a \E[\|\Delta U\|] \cdot L(U)^{1/2} \notag \\
    \leq & ~ \langle \nabla L(U), \E[ \Delta U ] \rangle + (b+\frac{a}{2 \eta}) \E[\|\Delta U\|^2] + \frac{a \eta}{2} L(U),
\end{align}
}
 where the second step follows from $2ab \leq ta^2 + \frac{1}{t}b^2$.

Plugging in $\Delta U : =-\frac{\tilde{\eta}}{K N} \sum_{k \in [K], c \in [N]} g_{c}\left(W_{c, k-1}\right)$, the first term in Eq.~\eqref{eq:nonconvex_one_round1} is
{%\small 
\begin{align}
    \label{eq:nonconvex_one_bound2}
    \langle \nabla L(U), \E[ \Delta U ] \rangle  
    = & ~ -\tilde{\eta} \langle \nabla L(U), \frac{1}{K N} \sum_{k \in [K], c \in [N]} \E[ g_{c}\left(W_{c, k-1}\right) ] \rangle  \notag \\
    = & ~ -\tilde{\eta} \langle \nabla L(U), \frac{1}{K N} \sum_{k \in [K], c \in [N]} \E[ \nabla L_{c}\left(W_{c, k-1}\right) ] \rangle  \notag \\
    \leq & ~ -\frac{\tilde{\eta}}{2} \|\nabla L(U)\|^2 + \frac{\tilde{\eta}}{2} \E \Big[ \Big\| \frac{1}{KN} \sum_{k \in K, c\in [N]} \nabla L_c(W_{c,k-1}) -\nabla L(U) \Big\|^2 \Big] \notag \\
    = & ~ -\frac{\tilde{\eta}}{2} \|\nabla L(U)\|^2 + \frac{\tilde{\eta}}{2} \E \Big[ \Big\| \frac{1}{KN} \sum_{k \in K, c\in [N]} ( \nabla L_c(W_{c,k-1}) -\nabla L_c(U) ) \Big\|^2 \Big]  \notag \\
    \leq & ~ -\frac{\tilde{\eta}}{2} \|\nabla L(U)\|^2  + \frac{\tilde{\eta}}{2KN} \sum_{k \in K, c\in [N]} \E[\|\nabla L_c(W_{c,k-1}) -\nabla L_c(U) \|^2] 
\end{align}
}
where the second step follows from $ \E[\Delta U]=-\frac{\tilde{\eta}}{K N} \sum_{k \in [K], c \in [N]} \E\left[\nabla L_{c}\left(W_{c, k-1}\right)\right]$, the third step follows from $-ab = \frac{1}{2}((b-a)^2-a^2) -\frac{1}{2}b^2 \leq \frac{1}{2}((b-a)^2-a^2)$, the fourth step follows from $L(U) := \frac{1}{N} \sum_{c=1}^N L_c(U)$, and the fifth step follows from $(\sum_{i=1}^n a_i)^2 \leq n\sum_{i=1}^n a_i^2$.

By Assumption 2 in the lemma statement and $2ab \leq a^2+b^2$, 
\begin{align}
    \label{eq:nonconvex_one_bound3}
     \|\nabla L_c(W_{c,k-1}) -\nabla L_c(U) \|^2  
    \leq & ~  \beta^2  \|W_{c,k-1} - U\|^2 + \alpha^2 \|W_{c,k-1} - U\|L_c(U)^{1/2} \notag \\
    \leq & ~\beta^2 \|W_{c,k-1} - U\|^2 + \frac{\alpha^2}{2} \|W_{c,k-1} - U\|^2 + \frac{\alpha^2}{2} L_c(U).
\end{align}
Combining (\ref{eq:nonconvex_one_bound2}) and (\ref{eq:nonconvex_one_bound3}), we have 
{%\small 
\begin{align}
    \label{eq:nonconvex_one_bound4}
     \langle \nabla L(U), \E[ \Delta U ] \rangle 
     \leq & ~ -\frac{\tilde{\eta}}{2} \|\nabla L(U)\|^2 + \frac{\tilde{\eta}}{2KN} \sum_{k \in K, c\in [N]} ((\beta^2+\frac{\alpha^2}{2}) \E[\|W_{c,k-1} - U\|^2]  + \frac{\alpha^2}{2}L_c(U)) \notag \\
     = & ~ -\frac{\tilde{\eta}}{2} \|\nabla L(U)\|^2 + \frac{\tilde{\eta}(2\beta^2+\alpha^2)}{4} \xi + \frac{\tilde{\eta}\alpha^2}{4} L(U).
\end{align}
}

Plugging in 
\begin{align*} 
\Delta U : =-\frac{\tilde{\eta}}{K N} \sum_{k \in [K], c \in [N]} g_{c}\left(W_{c, k-1}\right),
\end{align*}
we can upper bound the second term in (\ref{eq:nonconvex_one_round1}) as
{
\begin{align}
\label{eq:nonconvex_one_bound5}
  \|\Delta U\|^2 
    = & ~ \Big\| \frac{\tilde{\eta}}{K N} \sum_{k \in [K], c \in [N]} g_{c}\left(W_{c, k-1}\right) \Big\|^2 \notag \\
    \leq & ~ \Big\| \frac{\tilde{\eta}}{K N} \sum_{k \in [K], c \in [N]} \nabla L_{c}\left(W_{c, k-1}\right) - \tilde{\eta} \nabla L(U)  + \tilde{\eta} \nabla L(U) \Big\|^2 + \tilde{\eta}^2 \sigma^2  \notag \\
    \leq & ~ 2\tilde{\eta}^2 \Big\| \frac{1}{K N} \sum_{k \in [K], c \in [N]} (\nabla L_{c}\left(W_{c, k-1}\right) - \nabla L_c(U) )\Big\|^2  + 2\tilde{\eta}^2 \|  \nabla L(U) \|^2 + \tilde{\eta}^2 \sigma^2 \notag \\
    \leq & ~ \frac{2\tilde{\eta}^2}{KN}\sum_{k \in [K], c \in [N]}\|\nabla L_c(W_{c,k-1}) - L_c(U)\|^2 + 2\tilde{\eta}^2 \|  \nabla L(U) \|^2 + \tilde{\eta}^2 \sigma^2 \notag \\
    \leq & ~ \tilde{\eta}^2 (2 \beta^2 + \alpha^2) \xi + \tilde{\eta}^2 \alpha^2 L(U)  + 2\tilde{\eta}^2 \|  \nabla L(U) \|^2 + \tilde{\eta}^2 \sigma^2 .
\end{align}
}
where the first step follows from Lemma \ref{lem:smv}, the second step follows from $(a+b)^2 \leq 2a^2+2b^2$, the third step follows from $(\sum_{i=1}^n a_i)^2 \leq n\sum_{i=1}^n a_i^2$, and the last step follows from \eqref{eq:nonconvex_one_bound3}.

Combining (\ref{eq:nonconvex_one_round1}), (\ref{eq:nonconvex_one_bound4}) and (\ref{eq:nonconvex_one_bound5}), we upper bound the one round update as:{%\small
\begin{align*}
    & ~ \E[L(U+\Delta U)] - L(U)\\ 
    \leq & ~ -\frac{\tilde{\eta}}{2} \|\nabla L(U)\|^2 + \frac{\tilde{\eta}(2\beta^2+\alpha^2)}{2} \xi + \frac{\tilde{\eta}\alpha^2}{4} L(U)+(b+\frac{a}{2 \eta}) \\
     & ~ \cdot (\tilde{\eta}^2 (2 \beta^2 + \alpha^2) \xi + \tilde{\eta}^2 \alpha^2 L(U) + 2\tilde{\eta}^2 \|  \nabla L(U) \|^2 + \tilde{\eta}^2 \sigma^2)  + \frac{a \eta}{2} L(U) \\
     = & ~ -\frac{\tilde{\eta}}{2}(1-4b\tilde{\eta} - \frac{2a\tilde{\eta}}{\eta}) \|\nabla L(U)\|^2 + \tilde{\eta}^2(2 \beta^2 + \alpha^2)(\frac{1}{2\tilde{\eta}}+b+\frac{a}{2\eta})\xi + \tilde{\eta}^2 (b+\frac{a}{2\eta}) \sigma^2 \\
     & ~ + \Big(\frac{\tilde{\eta}\alpha^2}{4} + \tilde{\eta}^2 \alpha^2 (b+\frac{a}{2\eta}) + \frac{a \eta}{2} \Big) \cdot L(U).
\end{align*}
}

\end{proof}

\subsection{Proof of Theorem \ref{thm:main1_informal}}
\begin{proof}
Combining Lemma \ref{lem:xi_bound} and Lemma \ref{lem:nonconvex_one_round}, we get
\begin{align*}
    \E[L(U+\Delta U)] - L(U) 
    \leq & ~ -\frac{\tilde{\eta}}{2}(1-4b\tilde{\eta} - \frac{2a\tilde{\eta}}{\eta}) \|\nabla L(U)\|^2 \\
    & ~ + \tilde{\eta}^2 (b+\frac{a}{2\eta}) \sigma^2  + \tilde{\eta}^2(2 \beta^2 + \alpha^2)(\frac{1}{2\tilde{\eta}}+b+\frac{a}{2\eta})\xi  \\
     & ~ + \Big(\frac{\tilde{\eta}\alpha^2}{4} + \tilde{\eta}^2 \alpha^2 (b+\frac{a}{2\eta}) + \frac{a \eta}{2} \Big) \cdot L(U) \\
     \leq & ~ -\frac{\tilde{\eta}}{2}(1-4b\tilde{\eta} - \frac{2a\tilde{\eta}}{\eta}) \|\nabla L(U)\|^2 \\
     & ~ + \underbrace{ \tilde{\eta}^2 \Big( (2 \beta^2 + \alpha^2)(\frac{1}{2\tilde{\eta}}+b+\frac{a}{2\eta}) 20K\eta_l^2 +  (b+\frac{a}{2\eta}) \Big) }_{ {\cal A}_1 } \cdot \sigma^2 \\
     & ~ + \underbrace{ \tilde{\eta}^2(2 \beta^2 + \alpha^2)(\frac{1}{2\tilde{\eta}}+b+\frac{a}{2\eta})(20 \eta_l^2 + 40 K^2 \eta_l^2 \tau_2^2) }_{ {\cal A}_2 } \cdot L(U) \\
     & ~ + \underbrace{ \Big(\frac{\tilde{\eta}\alpha^2}{4} + \tilde{\eta}^2 \alpha^2 (b+\frac{a}{2\eta}) + \frac{a \eta}{2} \Big) }_{ {\cal A}_3 } \cdot L(U).
\end{align*}

By choosing $\eta = \wt{\eta}$ and $\eta_l = \min\{1/(\alpha^2 K), 1/(100KG(\beta+\alpha)), 1/(100\sqrt{K} (\beta + \alpha) ) \}$, we have
\begin{align*}
    {\cal A}_1 
    = & ~ \tilde{\eta}^2 \Big( (2 \beta^2 + \alpha^2)(\frac{1}{2\tilde{\eta}}+b+\frac{a}{2\eta}) 20K\eta_l^2 +  (b+\frac{a}{2\eta}) \Big) \\
    \leq & ~ (1+ a + b \wt{\eta}) \frac{\wt{\eta}}{10}
\end{align*}
and
\begin{align*}
     {\cal A}_2 
    = & ~ \tilde{\eta}^2(2 \beta^2 + \alpha^2)(\frac{1}{2\tilde{\eta}}+b+\frac{a}{2\eta})(20 \eta_l^2 + 40 K^2 \eta_l^2 \tau_2^2)~~~ \\
    \leq & ~ (1 + a + b \tilde{\eta}) \frac{\tilde{\eta}}{10},
\end{align*}
and also
\begin{align*}
   {\cal A}_3 
    = & ~ \Big(\frac{\tilde{\eta}\alpha^2}{4} + \tilde{\eta}^2 \alpha^2 (b+\frac{a}{2\eta}) + \frac{a \eta}{2} \Big)~~~~~~~~~~~~~~~~~~~~~~~~~~~ \\
    = & ~ \Big(\frac{\alpha^2 + 2a\alpha^2 + 2a}{4} + \alpha^2b\tilde{\eta} \Big) \tilde{\eta} \\
    \leq & ~ (\alpha^2 + a \alpha^2 + a + \alpha^2 b \wt{\eta}) \wt{\eta}.
\end{align*}
Then we can get
\begin{align}\label{eq:E_L_U_minus_L_U}
     \E[ L(U^{r}) ] - L(U^{r-1}) 
    \leq & ~ - \gamma_1 \| \nabla L(U^{r-1}) \|^2 + \gamma_2 L(U^{r-1}) + \gamma_3
\end{align}
where
\begin{align*}
    \gamma_1 : &= \frac{\tilde{\eta}}{2}(1-4b\tilde{\eta} - 2a), \\
    \gamma_2 : &= ( 1 + a + b \tilde{\eta}) \frac{\tilde{\eta}}{10} + ( \alpha^2 + a\alpha^2 + a + \alpha^2b\tilde{\eta})\tilde{\eta}, \\
    \gamma_3 : & = ( 1 + a + b \tilde{\eta}) \frac{\tilde{\eta}}{10} \sigma^2 .
\end{align*}
The Eq.~\eqref{eq:E_L_U_minus_L_U} can be written as follows:
\begin{align*}
    \E[ L(U^{r}) ] - L(U^*) 
    \leq & ~ ( L(U^{r-1}) - L(U^*) ) - \gamma_1 \| \nabla L(U^{r-1}) \|^2 + \gamma_2 L(U^{r-1}) + \gamma_3  \\
    \leq & ~  ( L(U^{r-1}) - L(U^*) )  -\gamma_1 \tau_1^2 L(U^{r-1})+  \gamma_2 L(U^{r-1}) + \gamma_3 \\
    \leq & ~  ( L(U^{r-1}) - L(U^*) ) - 0.5 \gamma_1 \tau_1^2 L(U^{r-1}) + \gamma_3 \\
    \leq & ~ (1-\gamma_4)   L(U^{r-1}) - L(U^*) + \gamma_3 \\
    \leq & ~ (1-\gamma_4)  ( L(U^{r-1}) - L(U^*) ) + \gamma_3.
\end{align*}
where the second step follows by Assumption 3 in the theorem statement, the third step follows from choosing $\gamma_2 \leq \gamma_1 \tau_1^2/2$ which means restricting $a \leq \tau_1^2/10$, $\alpha \leq \tau_1^2/10$ and $\tilde{\eta} \leq \tau_1^2/(20b)$, the fourth step follows from $\gamma_4 = 0.5 \gamma_1 \tau_1^2 = \frac{\tilde{\eta}}{4}(1-4b\tilde{\eta} - 2a)\tau_1^2$, and the last step follows from $L(U^*) \geq 0$.

Applying the above equation recursively,
\begin{align*}
    \E[L(U^{R})-L(U^*)]
    \leq & ~ (1-\gamma_4)^R \cdot (L(U^0) - L(U^*)) + 2\gamma_3
\end{align*}
\end{proof}

\subsection{Proof of Corollary \ref{coro:main1}}

\begin{proof}
By choosing 
\begin{align*} 
\eta_g \leq \min\{ \tau_1^2/(20Kb\eta_l), 2 \epsilon / (\sigma^2(1+a+\tau_1^2/20K)) \},
\end{align*}
we have $2\gamma_3 \leq \epsilon/2$. Then, we need to solve 
\begin{align*}
    (1-\gamma_4)^R \cdot (L(U^0) - L(U^*)) \leq \frac{\epsilon}{2},
\end{align*}
and by taking $\log$ function to solve for $R$, we need
\begin{align*}
    R \log (1-\gamma_4) \leq \log (\frac{\epsilon}{2(L(U^0) - L(U^*))}).
\end{align*}
Using the fact that $-1/\log(1-x) \leq 1/x$ for $0 < x \leq 1$, we have
\begin{align*}
    R \geq \log (\frac{2 (L(U^0) - L(U^*))}{\epsilon}) \frac{1}{\gamma_4}.
\end{align*}
\end{proof}

%% file: proof_sketch_app.tex
\section{Supplementary Proofs for Section~\ref{sec:proof_sketch}}\label{sec:proof_sketch_app}

\subsection{Proof for Proposition~\ref{prop:non_fed}}\label{sec:proof_non_fed_app}

\begin{proof}
We consider gradient update
 \begin{align*}
    x_{t+1} = x_t - \eta \cdot \nabla L(x) |_{x = x_t}.
 \end{align*}
We can compute $L(x_{t+1}) - L(x^*)$ by first applying $(a, b)$-semi-smoothness of $L$. {%\small
\begin{align}
    & ~ L(x_{t+1}) - L(x^*) \notag \\
    \leq & ~ L(x_t) - L(x^*) + \langle \nabla L(x_t), x_{t+1} - x_t \rangle 
      + b \| x_{t+1} - x_t \|^2 + a \| x_{t+1} - x_t \| \cdot L(x_t)^{1/2} \label{eq:non_fed1} \\
    = & ~ L(x_t) - L(x^*) - \eta \| \nabla L(x_t) \|^2 + b \eta^2 \| \nabla L(x_t) \|^2 + a \eta \| \nabla L(x_t) \| \cdot L(x_t)^{1/2} \notag \\
    := & ~ L(x_t) - L(x^*) + A \notag
\end{align}
}
By $(\tau_1, \tau_2)$-non-critical point, we know that
\begin{align*}
    A 
= & - \eta \| \nabla L(x_t) \|^2 +  b \eta^2 \| \nabla L(x_t) \|^2 
      + a \eta \| \nabla L(x_t) \| \cdot L(x_t)^{1/2} \\
     \le & ~ - \eta \tau_1^2 L(x_t) + b \eta^2 \tau_2^2 L(x_t) + a \eta \tau_2  L(x_t) \\
     = & ~ (- \eta \tau_1^2 +  b \eta^2 \tau_2^2 + a \eta \tau_2) L(x_t) \\
     := & ~ C \cdot L(x_t).
\end{align*}
Then, we have
\begin{align*}
    C ~=&~ - \eta \tau_1^2 +  b \eta^2 \tau_2^2 + a \eta \tau_2\\
    \le&~ - 0.5 \eta \tau_1^2 + b \eta^2 \tau_2^2 \\
    \le & ~ - 0.1 \eta \tau_1^2
\end{align*}
where the first step follows from $0.5 \tau_1^2 \geq a \tau_2$ and the second step follows from $\eta \leq \tau_1^2/(10b \tau_2^2)$.
Finally, 
\begin{align}
    L(x_{t+1}) - L(x^*) 
    \leq &~ (1 + C)L(x_t) - L(x^*) \notag\\
    \leq & ~ ( 1- 0.1 \eta \tau_1^2) \cdot L(x_t) - L(x^*) \notag \\
    \leq & ~ (1-\lambda) \cdot  L(x_t) - L(x^*)  \notag \\
    \leq  & ~ (1-\lambda) \cdot ( L(x_t) - L(x^*) ) \notag 
\end{align}

where the second last step follows $\lambda = 0.1 \eta \tau_1^2$. 

\end{proof}

%% file: discussion.tex
\section{Discussion and Conclusion}\label{sec:discussion}
 
In this paper, we analyze the convergence of \fedavg{} without the commonly-used smoothness assumption, and improve the theoretical convergence analysis of  \fedavg{} under the non-convex and non-smooth settings. Under the non-smooth setting, it is challenging to make suitable assumptions. We introduce the semi-smoothness assumption and non-critical point assumption to tackle these problems. Besides, when considering \fedavg{}, local drift is usually difficult to analyze and bound. Under our new theoretical framework, local drift and the progress of gradient are bounded appropriately. By our milder assumptions, we can prove the convergence of \fedavg{} with the vanilla SGD update.

Our work sheds light on the theoretical understanding of \fedavg{}. For future work, we hope our findings can provide insights for better FL algorithms design. In addition, in our paper, we show how to choose parameters for \fedavg, such as learning rate $\eta_g, \eta_l$, communication round $R$. We leave the detailed discussion about the effect of parameters in assumptions, like $\tau_1, \tau_2$, for future work. We can also try to further relax the assumptions. We hope our insights and proven techniques can inspire the following up works.

As a theoretical study, we would hardly expect its any direct detrimental societal consequences in the near future. We discuss the convergence of widely deployed FL algorithm \fedavg{} on more relaxed assumptions beyond smoothness. Although the results are promising,  
we should carefully examine the assumptions  
and aware the gap between theory and practice when using the theoretical results to guide algorithm deployment.

\section*{Acknowledgements}

The authors would like to thank Sen Li, Lianke Qin, Yitan Wang, Zheng Yu, Lichen Zhang, and Xiaofei Zhang for helpful discussions. The authors also like to thank anonymous reviewers for their helpful comments.

%% file: main.bbl
\newcommand{\etalchar}[1]{$^{#1}$}
\begin{thebibliography}{DMS{\etalchar{+}}21b}

\bibitem[AZLS19]{als19_dnn}
Zeyuan Allen-Zhu, Yuanzhi Li, and Zhao Song.
\newblock A convergence theory for deep learning via over-parameterization.
\newblock In {\em ICML}, pages 242--252. PMLR, 2019.

\bibitem[BDKD20]{bdkd20}
Debraj Basu, Deepesh Data, Can Karakus, and Suhas~N Diggavi.
\newblock Qsparse-local-sgd: Distributed sgd with quantization, sparsification,
  and local computations.
\newblock {\em IEEE Journal on Selected Areas in Information Theory},
  1(1):217--226, 2020.

\bibitem[BLO05]{blo05}
James~V Burke, Adrian~S Lewis, and Michael~L Overton.
\newblock A robust gradient sampling algorithm for nonsmooth, nonconvex
  optimization.
\newblock {\em SIAM Journal on Optimization}, 15(3):751--779, 2005.

\bibitem[CMOB19]{cmo+19}
Mingqing Chen, Rajiv Mathews, Tom Ouyang, and Fran{\c{c}}oise Beaufays.
\newblock Federated learning of out-of-vocabulary words.
\newblock {\em arXiv preprint arXiv:1903.10635}, 2019.

\bibitem[DMS{\etalchar{+}}21a]{dms+21}
Wei Deng, Yi-An Ma, Zhao Song, Qian Zhang, and Guang Lin.
\newblock On convergence of federated averaging langevin dynamics.
\newblock {\em arXiv preprint arXiv:2112.05120}, 2021.

\bibitem[DMS{\etalchar{+}}21b]{dmszl21}
Wei Deng, Yi-An Ma, Zhao Song, Qian Zhang, and Guang Lin.
\newblock On convergence of federated averaging langevin dynamics.
\newblock {\em arXiv preprint arXiv:2112.05120}, 2021.

\bibitem[GHR21]{ghr21}
Eduard Gorbunov, Filip Hanzely, and Peter Richt{\'a}rik.
\newblock Local sgd: Unified theory and new efficient methods.
\newblock In {\em International Conference on Artificial Intelligence and
  Statistics}, pages 3556--3564. PMLR, 2021.

\bibitem[HKMM20]{hkmm20}
Farzin Haddadpour, Mohammad~Mahdi Kamani, Aryan Mokhtari, and Mehrdad Mahdavi.
\newblock Federated learning with compression: Unified analysis and sharp
  guarantees.
\newblock {\em arXiv preprint arXiv:2007.01154}, 2020.

\bibitem[HLSY21]{hlsy21}
Baihe Huang, Xiaoxiao Li, Zhao Song, and Xin Yang.
\newblock Fl-ntk: A neural tangent kernel-based framework for federated
  learning analysis.
\newblock In {\em ICML}, pages 4423--4434. PMLR, 2021.

\bibitem[HRM{\etalchar{+}}18]{hrm+18}
Andrew Hard, Kanishka Rao, Rajiv Mathews, Fran{\c{c}}oise Beaufays, Sean
  Augenstein, Hubert Eichner, Chlo{\'e} Kiddon, and Daniel Ramage.
\newblock Federated learning for mobile keyboard prediction.
\newblock {\em arXiv preprint arXiv:1811.03604}, 2018.

\bibitem[KKM{\etalchar{+}}20]{kkm+20}
Sai~Praneeth Karimireddy, Satyen Kale, Mehryar Mohri, Sashank Reddi, Sebastian
  Stich, and Ananda~Theertha Suresh.
\newblock Scaffold: Stochastic controlled averaging for federated learning.
\newblock In {\em ICML}, pages 5132--5143. PMLR, 2020.

\bibitem[KLB{\etalchar{+}}20]{klb+20}
Anastasia Koloskova, Nicolas Loizou, Sadra Boreiri, Martin Jaggi, and Sebastian
  Stich.
\newblock A unified theory of decentralized sgd with changing topology and
  local updates.
\newblock In {\em ICML}, pages 5381--5393. PMLR, 2020.

\bibitem[KMA{\etalchar{+}}19]{kma+19}
Peter Kairouz, H~Brendan McMahan, Brendan Avent, Aur{\'e}lien Bellet, Mehdi
  Bennis, Arjun~Nitin Bhagoji, Keith Bonawitz, Zachary Charles, Graham Cormode,
  Rachel Cummings, et~al.
\newblock Advances and open problems in federated learning.
\newblock {\em arXiv preprint arXiv:1912.04977}, 2019.

\bibitem[KMR19]{kmr19}
Ahmed Khaled, Konstantin Mishchenko, and Peter Richt{\'a}rik.
\newblock First analysis of local gd on heterogeneous data.
\newblock {\em arXiv preprint arXiv:1909.04715}, 2019.

\bibitem[KMR20]{kmr20}
Ahmed Khaled, Konstantin Mishchenko, and Peter Richt{\'a}rik.
\newblock Tighter theory for local {SGD} on indentical and heterogeneous data.
\newblock In {\em Proceedings of AISTATS}, 2020.

\bibitem[KMRR16]{kmrr16}
Jakub Kone{\v{c}}n{\`y}, H~Brendan McMahan, Daniel Ramage, and Peter
  Richt{\'a}rik.
\newblock Federated optimization: Distributed machine learning for on-device
  intelligence.
\newblock {\em arXiv preprint arXiv:1610.02527}, 2016.

\bibitem[KRSJ19]{krsj19}
Sai~Praneeth Karimireddy, Quentin Rebjock, Sebastian~U Stich, and Martin Jaggi.
\newblock Error feedback fixes signsgd and other gradient compression schemes.
\newblock {\em arXiv preprint arXiv:1901.09847}, 2019.

\bibitem[LGD{\etalchar{+}}20]{lgd+20}
Xiaoxiao Li, Yufeng Gu, Nicha Dvornek, Lawrence Staib, Pamela Ventola, and
  James~S Duncan.
\newblock Multi-site fmri analysis using privacy-preserving federated learning
  and domain adaptation: Abide results.
\newblock {\em arXiv preprint arXiv:2001.05647}, 2020.

\bibitem[LHY{\etalchar{+}}19]{lhy+19}
Xiang Li, Kaixuan Huang, Wenhao Yang, Shusen Wang, and Zhihua Zhang.
\newblock On the convergence of fedavg on non-iid data.
\newblock {\em arXiv preprint arXiv:1907.02189}, 2019.

\bibitem[LJZ{\etalchar{+}}21]{ljz+21}
Xiaoxiao Li, Meirui Jiang, Xiaofei Zhang, Michael Kamp, and Qi~Dou.
\newblock Fed{BN}: Federated learning on non-{IID} features via local batch
  normalization.
\newblock In {\em ICLR}, 2021.

\bibitem[LSTS20]{lst+20}
Tian Li, Anit~Kumar Sahu, Ameet Talwalkar, and Virginia Smith.
\newblock Federated learning: Challenges, methods, and future directions.
\newblock {\em IEEE Signal Processing Magazine}, 37(3):50--60, 2020.

\bibitem[LSZ{\etalchar{+}}20]{lsz+20}
Tian Li, Anit~Kumar Sahu, Manzil Zaheer, Maziar Sanjabi, Ameet Talwalkar, and
  Virginia Smith.
\newblock Federated optimization in heterogeneous networks.
\newblock In {\em Conference on Machine Learning and Systems, 2020a}, 2020.

\bibitem[LY17]{ly17}
Yuanzhi Li and Yang Yuan.
\newblock Convergence analysis of two-layer neural networks with {R}e{LU}
  activation.
\newblock In {\em NeurIPS}, pages 597--607, 2017.

\bibitem[MMR{\etalchar{+}}17]{mmr+17}
Brendan McMahan, Eider Moore, Daniel Ramage, Seth Hampson, and Blaise~Aguera
  y~Arcas.
\newblock Communication-efficient learning of deep networks from decentralized
  data.
\newblock In {\em Artificial Intelligence and Statistics}, pages 1273--1282.
  PMLR, 2017.

\bibitem[PD19]{pd19}
Kumar~Kshitij Patel and Aymeric Dieuleveut.
\newblock Communication trade-offs for synchronized distributed {SGD} with
  large step size.
\newblock {\em arXiv preprint arXiv:1904.11325}, 2019.

\bibitem[RMRB19]{rmr19+}
Swaroop Ramaswamy, Rajiv Mathews, Kanishka Rao, and Fran{\c{c}}oise Beaufays.
\newblock Federated learning for emoji prediction in a mobile keyboard.
\newblock {\em arXiv preprint arXiv:1906.04329}, 2019.

\bibitem[SK19]{sk19}
Sebastian~U Stich and Sai~Praneeth Karimireddy.
\newblock The error-feedback framework: Better rates for sgd with delayed
  gradients and compressed communication.
\newblock {\em arXiv preprint arXiv:1909.05350}, 2019.

\bibitem[Sti18]{s18}
Sebastian~U Stich.
\newblock Local sgd converges fast and communicates little.
\newblock {\em arXiv preprint arXiv:1805.09767}, 2018.

\bibitem[SWYZ22]{swyz22}
Zhao Song, Yitan Wang, Zheng Yu, and Lichen Zhang.
\newblock Sketching for first order method: Efficient algorithm for
  low-bandwidth channel and vulnerability.
\newblock {\em arXiv preprint arXiv:2210.08371}, 2022.

\bibitem[SYZ21]{syz22_iclr}
Zhao Song, Zheng Yu, and Lichen Zhang.
\newblock Iterative sketching and its application to federated learning.
\newblock {\em openreview}, 2021.

\bibitem[WTS{\etalchar{+}}19]{wts+19}
Shiqiang Wang, Tiffany Tuor, Theodoros Salonidis, Kin~K. Leung, Christian
  Makaya, Ting He, and Kevin Chan.
\newblock Adaptive federated learning in resource constrained edge computing
  systems.
\newblock {\em IEEE Journal on Selected Areas in Communications},
  37(6):1205--1221, 2019.

\bibitem[WYS{\etalchar{+}}20]{wys+20}
Hongyi Wang, Mikhail Yurochkin, Yuekai Sun, Dimitris Papailiopoulos, and
  Yasaman Khazaeni.
\newblock Federated learning with matched averaging.
\newblock {\em arXiv preprint arXiv:2002.06440}, 2020.

\bibitem[YAE{\etalchar{+}}18]{yae+18}
Timothy Yang, Galen Andrew, Hubert Eichner, Haicheng Sun, Wei Li, Nicholas
  Kong, Daniel Ramage, and Fran{\c{c}}oise Beaufays.
\newblock Applied federated learning: Improving google keyboard query
  suggestions.
\newblock {\em arXiv preprint arXiv:1812.02903}, 2018.

\bibitem[YHW{\etalchar{+}}19]{yhwz+19}
Xin Yao, Tianchi Huang, Chenglei Wu, Rui-Xiao Zhang, and Lifeng Sun.
\newblock Federated learning with additional mechanisms on clients to reduce
  communication costs.
\newblock {\em arXiv preprint arXiv:1908.05891}, 2019.

\bibitem[YYZ19]{yyz19}
Hao Yu, Sen Yang, and Shenghuo Zhu.
\newblock Parallel restarted {SGD} with faster convergence and less
  communication: Demystifying why model averaging works for deep learning.
\newblock In {\em Proceedings of the AAAI Conference on Artificial
  Intelligence}, volume~33, pages 5693--5700, 2019.

\bibitem[ZLL{\etalchar{+}}18]{zll18}
Yue Zhao, Meng Li, Liangzhen Lai, Naveen Suda, Damon Civin, and Vikas Chandra.
\newblock Federated learning with non-iid data.
\newblock {\em arXiv preprint arXiv:1806.00582}, 2018.

\bibitem[ZWLS10]{zwls10}
Martin Zinkevich, Markus Weimer, Lihong Li, and Alex~J Smola.
\newblock Parallelized stochastic gradient descent.
\newblock In {\em Advances in neural information processing systems}, pages
  2595--2603, 2010.

\end{thebibliography}
